\documentclass[letterpaper]{article}
\usepackage[preprint]{aaai2027}
\usepackage[hyphens]{url}
\usepackage{graphicx}
\usepackage{natbib}
\usepackage{caption}
\usepackage{amsmath}
\usepackage{amssymb}
\usepackage{amsthm}
\usepackage{algorithm}
\usepackage{algorithmic}
\usepackage{booktabs}
\usepackage{multirow}

\newcommand{\sd}[1]{{\scriptsize$\pm$#1}}

\newcommand{\cmark}{\checkmark}
\newcommand{\xmark}{\ensuremath{\times}}

\title{H\textsuperscript{2}EDL: Hyper Evidential Deep Learning for Hierarchical Classification}
\author{
Yuanye Liu\textsuperscript{\rm 1},
Xiahai Zhuang\textsuperscript{\rm 1}
}
\affiliations{
\textsuperscript{\rm 1} School of Data Science, Fudan University
}

\begin{document}
\maketitle

\begin{abstract}
Fine-grained recognition often involves hierarchical label spaces, where a model may be confident about a coarse semantic concept while remaining uncertain among its descendant classes.
Such structured ambiguity requires uncertainty representations that capture both fine-grained classes and intermediate concepts.
However, existing tools each capture only half of it: flat evidential classifiers quantify total ignorance with a single vacuity on the leaf frame, and hierarchical classifiers propagate point probabilities with no notion of evidence.
Hyper-opinions would unify the two, but their general form is exponential in the label count, and existing hyper-evidential networks either require composite labels to be supplied in the training data or read them off an unstructured weight pattern, with no principled notion of which composites deserve mass.
We observe that the taxonomy itself is the missing hyperdomain. Its subtrees and leaf singletons form a linear-size focal family, and one local Dirichlet opinion per branching node induces every composite mass in closed form.
The resulting model, H$^2$EDL, can be interpreted in two complementary ways using the same set of parameters. From a prediction perspective, it functions as a hierarchical classifier that preserves consistency across different levels of the label tree. From a probabilistic perspective, it defines a valid tree-structured hyper-opinion, where the mass assigned to each node represents the belief that reaches that node but does not provide sufficient confidence to further specialize into its descendants.
On FGVC-Aircraft and DERM12345, H$^2$EDL reduces calibration error by approximately half compared with cross-entropy baselines, with the improvement becoming more pronounced at deeper hierarchy levels and under larger training budgets. Despite achieving similar leaf-level accuracy, it preserves the correct coarse category 19\% more often when making fine-grained mistakes.
Code will be released.

\end{abstract}

\section{Introduction}

\begin{figure}[t]
    \centering
    \includegraphics[width=\columnwidth]{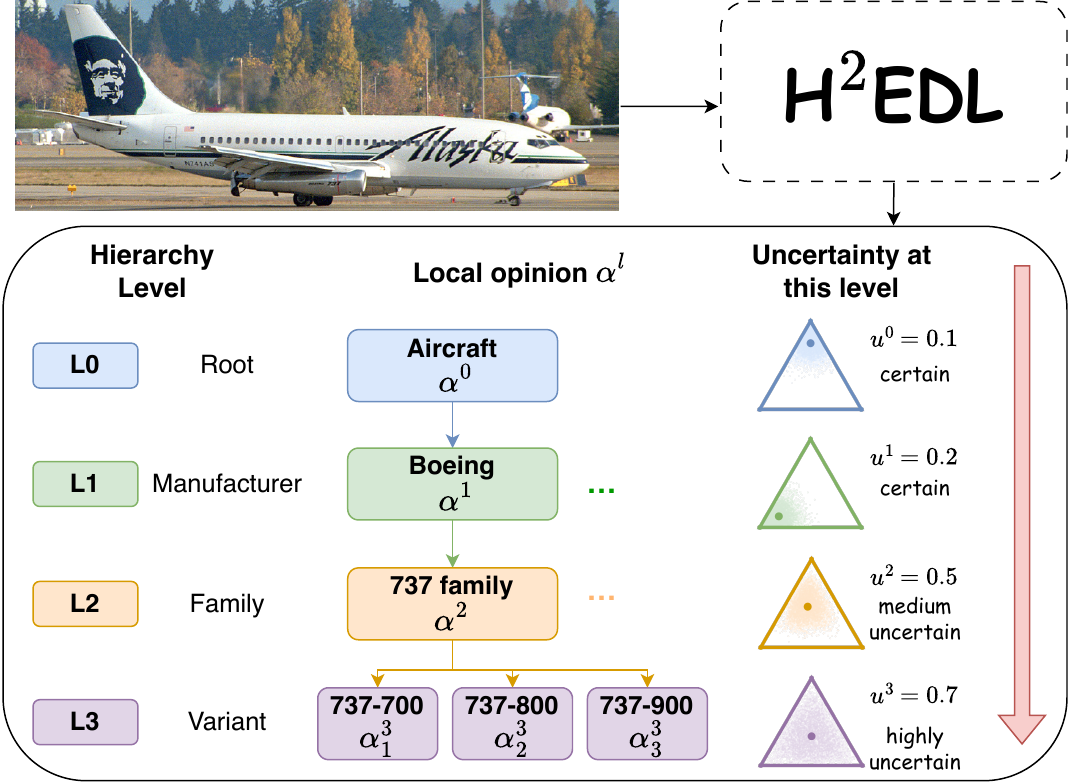}
    \caption{
    Uncertainty on a label hierarchy has a location. H$^2$EDL carries a local opinion at every branching decision, so it reports a separate vacuity for each level: the evidence is decisive about the manufacturer ($u^1 = 0.2$), already thinner about the family ($u^2 = 0.5$), and runs out among the $737$ variants ($u^3 = 0.7$), where a flat evidential model has one vacuity for the entire leaf level.
    }
    \label{fig:teaser}
    \vspace{-1.5em}
\end{figure}

Uncertainty-aware classification asks a model to report not only a label but how much its prediction can be trusted, and a substantial literature now supplies that number through calibration, Bayesian approximation, or evidential parameterizations \citep{guo2017calibration, gal2016dropout, lakshminarayanan2017ensembles, sensoy2018edl}. When the label space is a hierarchy, however, a single number cannot carry that message.
Consider a model that classifies aircraft assigning $0.4$ to the Boeing 737-800 (Fig.~\ref{fig:teaser}). It may be in either of two very different states: it may have settled that the aircraft is a 737 and be undecided only among the -700, -800 and -900 variants, or it may be spreading the remaining mass across unrelated manufacturers. The confidence is identical; the two predictions are not. What separates them is not how much uncertainty the model holds but where that uncertainty sits.

Flat evidential deep learning (EDL)~\citep{sensoy2018edl} can estimate the uncertainty, but it attaches to the leaf level only, where one global scalar cannot depict hierarchy.
Hierarchical classifiers~\citep{silla2011survey,bertinetto2020making} incorporate the label tree into prediction, but typically still rely on flat probability normalization within local decisions.
Subjective Logic supplies the object that
would unify them, the \emph{hyper-opinion}~\citep{josang2016sl}, which assigns
mass to composite subsets.
Its general form admits up to $2^{L}-1$ focal sets.
Existing hyper-evidential networks rely on composite labels that must be {manually specified in the training set}, despite the absence of an inherent structure to define such labels~\citep{li2024henn}, or by letting the composites fall out of the network's own weight structure \citep{qu2024hedl}. Both operate on a flat frame, which offers no principled way to say which composites deserve mass.

A taxonomy naturally provides the required hyperdomain. The sets of descendant leaves associated with internal nodes, together with leaf singletons, define a focal family with size at most $L+|\mathcal{B}|$, which grows linearly with the tree structure rather than exponentially with the number of labels. Moreover, each element of this family corresponds to a meaningful concept already represented in the taxonomy.
Additionally, a local Dirichlet opinion at each branching node determines the complete hyper-opinion in closed form. Standard leaf supervision naturally trains these local opinions, since predicting a leaf label is equivalent to making a sequence of decisions along the tree. Thus, the taxonomy simultaneously provides the hyperdomain, the parameterization, and the supervision signal. No additional label construction, auxiliary prediction network, or post-hoc calibration procedure is required.

Concretely, H\textsuperscript{2}EDL places an evidential head at each branching node and computes leaf probabilities by multiplying conditional means along root-to-leaf paths. The same concentration parameters admit two complementary interpretations. Predictively, they define a normalized and hierarchically consistent leaf classifier, from which predictions at any coarse level can be obtained directly. From an uncertainty perspective, they induce a tree-structured hyper-opinion, where the mass at node \(v\) represents belief that reaches \(v\) but lacks sufficient evidence to descend further.

We prove that this mass assignment is valid, that the predictive distribution lies within the belief--plausibility bounds of the induced hyper-opinion for every subset of leaves, and that non-specific mass can be decomposed exactly across hierarchy depths. The model can therefore identify where evidence becomes insufficient along each prediction path without additional inference procedures.

Unlike flat models that compare all \(L\) leaf classes at once, H\textsuperscript{2}EDL makes a sequence of local decisions. Each decision is trained with all samples in its subtree, so uncertainty should increase gradually at deeper levels. Ancestor decisions also restrict the possible error region, encouraging errors to remain within nearby subtrees. We evaluate both properties experimentally. We first examine whether existing benchmarks can distinguish hierarchical classifiers and find that their leaf accuracy is often dominated by a single coarse decision, which can obscure gains from fine-grained hierarchical modeling.

\smallskip\noindent\textbf{Contributions.} (i)~A tractable hyper-evidential
formulation, in which a taxonomy supplies a linear-size focal family on which
composite belief is induced rather than predicted. This removes the
composite-label requirement that has confined prior hyper-evidential networks to
prescribed label groups. (ii)~An operational classifier that is hierarchically
coherent by construction and exposes a vacuity at every individual decision.
(iii)~Theory linking the two readings: validity of the induced hyper-opinion
(Prop.~1), containment of the predictor in its credal set (Prop.~2), and an
exact depth decomposition of non-specific mass (Prop.~3). (iv)~An evaluation matched to the problem. Having established that
leaf accuracy measures one coarse decision common to all methods, we score every
level, and find H\textsuperscript{2}EDL ahead of all baselines at both
intermediate DERM12345 levels.

\section{Related Work}
\paragraph{Hierarchical classification.} Label
taxonomies~\citep{silla2011survey} are exploited by encoding the tree in the loss
or in embedding geometry~\citep{bertinetto2020making}, by minimizing a
hierarchical risk post hoc over an unchanged probability
vector~\citep{karthik2021crm}, or by hedging a prediction upward to an
ancestor~\citep{deng2012hedging}. All of these act on point
probabilities. The tree constrains or rescores a distribution that contains no
notion of evidence, so ambiguity confined to one subtree and ambiguity spread
across the frame remain the same object. Our aim is to give the prediction a
structured uncertainty that the tree makes both meaningful and cheap.

\paragraph{Evidential deep learning.} EDL~\citep{sensoy2018edl} predicts
Dirichlet concentrations and obtains closed-form epistemic uncertainty in a
single forward pass. Prior networks~\citep{malinin2018dpn}, the family surveyed
by \citet{ulmer2023survey}, and refinements that relax the prior weight and
variance term~\citep{chen2024redl} or reweight by Fisher
information~\citep{deng2023iedl} all share that design, and all keep a flat
frame with a single vacuity for the entire label set. Closest to us,
HENN \citep{li2024henn} reaches hyper-opinions through grouped Dirichlets over a prescribed collection of composite labels, which must be given in the training data; HEDL \citep{qu2024hedl} instead lets the composites emerge from the sign pattern of the final linear layer, and then projects the hyper-opinion back to a flat multinomial opinion for training and prediction.
Both methods operate on an unstructured class frame, where no mechanism determines which composite hypotheses should receive mass. Consequently, composite masses merely reflect uncertainty over arbitrary class subsets without revealing what those subsets represent.

\providecommand{\sd}[1]{{\scriptsize\,$\pm$#1}}

\section{Method}
\label{sec:method}

H\textsuperscript{2}EDL assigns every branching node of the taxonomy a local
Dirichlet opinion over its children and multiplies these opinions along
root-to-leaf paths. One parameter set then supports two readings: an operational
classifier whose levels cannot disagree, and a tree-structured hyper-opinion in
which composite mass is induced rather than predicted. After fixing notation we
develop the two readings in turn.

\subsection{Notation and Evidential Background}
We classify over a label hierarchy given as a rooted tree $\mathcal{T}$ with root
$r$. Leaves $\mathcal{L}$ ($|\mathcal{L}|=L$) are the fine-grained targets;
internal nodes are coarser concepts. Node $v$ has children $\mathcal{C}(v)$,
$K_v=|\mathcal{C}(v)|$; it is \emph{branching} if $K_v\ge2$ and
\emph{pass-through} if $K_v=1$, and $\mathcal{B}$ denotes the branching nodes.
Each leaf $\ell$ has a unique root-to-leaf path $\pi(\ell)$ inducing a set of
\emph{branching decisions}
$\mathcal{D}(\ell)=\{(v,c_v):v\in\mathcal{B}\cap\pi(\ell)\}$, where $c_v$ indexes
the child of $v$ on the path; pass-through nodes carry no decision.

Evidential Deep Learning~\citep{sensoy2018edl} treats a $K$-way prediction as a
Dirichlet over the simplex: the network emits non-negative evidence
$e\in\mathbb{R}^K_{\ge0}$ and forms $\boldsymbol{\alpha}=e+W\boldsymbol{a}$ with
prior weight $W>0$, base rate $\boldsymbol{a}\in\Delta^{K-1}$ and strength
$S=\sum_k\alpha_k$. In Subjective Logic~\citep{josang2016sl} this is an
\emph{opinion} $(b,u,a)$ with
\begin{equation}
b_k=\frac{e_k}{S},\qquad
u=\frac{W}{S}\in(0,1],\qquad
\hat p_k=\frac{\alpha_k}{S}=b_k+a_k u,
\label{eq:opinion}
\end{equation}
and $\sum_k b_k+u=1$. Vacuity $u$ is large when total evidence is small: the
model ``does not know.'' Sensoy EDL is the case $W\!=\!K$, $a_k\!=\!1/K$. We
exploit vacuity per node to reason about granularity.

\subsection{Operational Hierarchical Classifier}
Rather than one Dirichlet over $L$ leaves, H\textsuperscript{2}EDL attaches a
lightweight evidential head to every branching node. From the shared
backbone feature $f=\phi(x)$ each head produces a local opinion over the $K_v$
children,
\begin{equation}
e_v=\operatorname{softplus}\!\bigl(W_v^{\top}f+b_v\bigr),
\qquad
\boldsymbol{\alpha}_v=e_v+W\,a_v,
\label{eq:head}
\end{equation}

with $b_{v,k},u_v,\hat p_{v,k}$ following Eq.~\eqref{eq:opinion}. Crucially
$\hat{\boldsymbol{p}}_v$ is read as the conditional $P(\text{child}\mid
v,x)$. Setting $a_v=\frac1{K_v}\mathbf{1}$, $W=K_v$ recovers the standard EDL
prior. For imbalanced children we also allow a \emph{tempered inverse-frequency}
base rate, $a_{v,k}\propto(n_{v,k}+s)^{-\tau}$ over the training counts $n_{v,k}$
with smoothing $s$, temperature $\tau\in[0,1]$ ($\tau\!=\!0$ uniform) and
$W=w_{\mathrm{scale}}K_v$: rarer children start from a higher baseline belief, so
less evidence is needed to predict them. It is a fixed buffer adding no
parameters; we refer to it as the \emph{base-rate} variant. Pass-through
nodes receive neither head nor loss. Local opinions compose into a leaf distribution by
multiplying projected means along each path,
\begin{equation}
P(\ell\mid x)=\!\!\prod_{(v,c)\in\mathcal{D}(\ell)}\!\!\hat p_{v,c}
\;=\!\!\prod_{(v,c)\in\mathcal{D}(\ell)}\!\!\frac{\alpha_{v,c}}{S_v},
\label{eq:leafprob}
\end{equation}
and the probability of any internal concept is recovered exactly by the
telescoping partial product,
\begin{equation}
P(v\mid x)=\!\!\sum_{\ell\in\mathrm{leaves}(v)}\!\!P(\ell\mid x)
=\!\!\prod_{(u,c)\in\mathcal{D}(v)}\!\!\hat p_{u,c}.
\label{eq:subtree}
\end{equation}

\noindent\textbf{Proposition 0} (normalization \& coherence)\textbf{.} For any
local means $\hat{\boldsymbol{p}}_v\in\Delta^{K_v-1}$: (i)~$\sum_{\ell}P(\ell\mid
x)=1$; (ii)~coarse probabilities equal Eq.~\eqref{eq:subtree}; (iii)~if $v$ is an
ancestor of $v'$ then $P(v'\mid x)\le P(v\mid x)$.

\smallskip\noindent Coarse-level evaluation therefore needs no separate head, and
the model is hierarchically coherent by construction. We show next
that $\{\boldsymbol{\alpha}_v\}$ carries strictly more structure than the point
distribution of Eq.~\eqref{eq:leafprob}.

\subsection{The Induced Tree-Structured Hyper-Opinion}
\label{sec:hyperopinion}
A hyper-opinion~\citep{josang2016sl} assigns mass to singleton and
composite subsets of a frame; a leaf frame admits $2^L-1$ of them, impractical to
parameterize or supervise. H\textsuperscript{2}EDL instead lets the taxonomy
select the focal family
\begin{equation}
\mathcal{F}_{\mathcal{T}}=
\bigl\{\{\ell\}:\ell\in\mathcal{L}\bigr\}\cup
\bigl\{\mathrm{leaves}(v):v\in\mathcal{B}\bigr\},
\label{eq:hyperdomain}
\end{equation}
so $|\mathcal{F}_{\mathcal{T}}|\le L+|\mathcal{B}|$ is linear in the tree and
every focal set is a named concept. Prior evidential work requires composite
sets to be given as training labels on a flat frame~\citep{li2024henn};
here all masses are induced from ordinary leaf supervision, with zero mass
outside Eq.~\eqref{eq:hyperdomain}. Writing each local opinion in belief form and
defining the \emph{reach mass}
$m_v=\prod_{(w,c)\in\mathcal{D}(v)}b_{w,c}$ (so $m_r=1$), assign
\begin{equation}
m(\{\ell\})=\!\!\prod_{(v,c)\in\mathcal{D}(\ell)}\!\! b_{v,c},
\qquad
m\bigl(\mathrm{leaves}(v)\bigr)=m_v\,u_v .
\label{eq:hypermass}
\end{equation}
Each subtree mass $m_vu_v$ is the share of belief that reaches $v$ but
cannot be committed to any child; at the root $m(\mathcal{L})=u_r$. Pass-through
nodes create no focal set, having no uncertain decision. Note the operational
predictor multiplies means, whereas singleton mass here is a product of
beliefs; the latter are smaller and need not sum to one over leaves.

\smallskip\noindent\textbf{Proposition 1} (validity)\textbf{.} All masses in
Eq.~\eqref{eq:hypermass} are nonnegative and $\sum_{\ell}m(\{\ell\})+
\sum_{v\in\mathcal{B}}m_vu_v=1$; i.e.\ $\{\boldsymbol{\alpha}_v\}$ induces a
valid, tree-supported belief mass assignment, which is a sparse
Subjective-Logic hyper-opinion on $\mathcal{F}_{\mathcal{T}}$.

\smallskip\noindent\textbf{Proposition 2} (credal containment)\textbf{.} With
$\mathrm{bel}(A)=\!\sum_{F\subseteq A}m(F)$ and
$\mathrm{pl}(A)=\!\sum_{F\cap A\ne\varnothing}m(F)$, both summed over
$F\in\mathcal{F}_{\mathcal{T}}$, the H\textsuperscript{2}EDL distribution
satisfies $\mathrm{bel}(A)\le P(A\mid x)\le\mathrm{pl}(A)$ for \emph{every}
$A\subseteq\mathcal{L}$: it lies in the credal set of the induced hyper-opinion.
Taking $A=\{\ell\}$ gives $\mathrm{bel}(\ell)=m(\{\ell\})$ and
$\mathrm{pl}(\ell)=m(\{\ell\})+\sum_{v:\,\ell\in\mathrm{leaves}(v)}m_vu_v$.

\smallskip\noindent\textbf{Proposition 3} (depth decomposition)\textbf{.} The
total non-specific mass $U=1-\sum_\ell\mathrm{bel}(\ell)$ decomposes exactly by
depth: $U=\sum_d U_d$ with
$U_d=\sum_{v\in\mathcal{B}:\,d(v)=d}m_vu_v$.

\smallskip\noindent Proposition~2 turns Eq.~\eqref{eq:leafprob} into a
projection of the hyper-opinion: the per-leaf imprecision band
$\mathrm{pl}(\ell)-\mathrm{bel}(\ell)=\sum_{v\ni\ell}m_vu_v$ is driven by
ancestral vacuities weighted by reach mass, where a flat evidential model has one
global width $u$. Proposition~3 gives a depth-resolved profile: $U_d$ is the mass
that refuses to descend past depth $d$, computed per sample at no extra cost.
It is the formal counterpart of ``how far down the tree did the evidence
carry?'', and we treat it as a structural decomposition of where belief stops
rather than a calibrated epistemic-uncertainty estimate.

\subsection{Hierarchical Evidential Objective}
We supervise only the branching nodes on each sample's ground-truth path,
mirroring the conditional factorization of Eq.~\eqref{eq:leafprob}. For node $v$
with one-hot child target $y_v$ we use the Bayes-risk EDL
loss~\citep{sensoy2018edl}, $\mathcal{L}^{\mathrm{edl}}_v=
\sum_k(y_{v,k}-\hat p_{v,k})^2+\sum_k\hat p_{v,k}(1-\hat p_{v,k})/(S_v+1)$. To
suppress evidence on wrong children we add an annealed KL term that pulls
the off-target Dirichlet toward the prior: masking the target class to its prior
value, $\tilde{\boldsymbol{\alpha}}_v=y_v\odot(Wa_v)+(1-y_v)\odot
\boldsymbol{\alpha}_v$, we penalize
$\mathcal{L}^{\mathrm{kl}}_v=\mathrm{KL}(\mathrm{Dir}(\tilde{\boldsymbol{\alpha}}_v)
\|\mathrm{Dir}(Wa_v))$, which reduces to the standard Sensoy regularizer for
uniform $a_v$. The per-sample hierarchical loss sums over the ground-truth path
with depth weights $w_d$ (we use $w_d\!=\!1$):
\begin{equation}
\mathcal{L}^{\mathrm{path}}=
\!\!\sum_{(v,c)\in\mathcal{D}(\ell^\star)}\!\!
w_{d(v)}\bigl(\mathcal{L}^{\mathrm{edl}}_v+\lambda_t\,\mathcal{L}^{\mathrm{kl}}_v\bigr),
\label{eq:path}
\end{equation}
with $\lambda_t$ annealed from $0$ to $\lambda_{\max}$ so evidence can accumulate
before regularization. We additionally optimize the quantity the model is
evaluated on by adding a negative log-likelihood on the differentiable path
product, $\mathcal{L}^{\mathrm{leaf}}=-\log P(\ell^\star\mid x)=
\sum_{(v,c)\in\mathcal{D}(\ell^\star)}-\log\hat p_{v,c}$. The identity makes its
role precise: the leaf NLL is separable into per-node log-losses, so it adds no
interaction between heads beyond the shared backbone, only a cross-entropy on
each on-path $\hat p_{v,c}$ alongside the Bayes-risk term. The objective is
$\mathcal{L}=\mathcal{L}^{\mathrm{path}}+\beta\mathcal{L}^{\mathrm{leaf}}$ with
$\beta\!=\!0.1$. An optional tree-consistency regularizer, penalizing vacuity
that increases from a parent to a confidently selected child, is referred
to as the \emph{consistency} variant.

\paragraph{Partial hierarchical supervision.} Because supervision enters
Eq.~\eqref{eq:path} as a set of decisions rather than a leaf index, an
annotation naming only an ancestor $v^\star$ is not a degraded label but a
shorter one: the sum runs over $\mathcal{D}(v^\star)$. No architectural or
objective change is needed, and heads below $v^\star$ stay free to express
vacuity rather than being taught a fabricated target. A flat model, by contrast,
must either discard the sample or invent a distribution over $v^\star$'s
descendants.

\subsection{What the Factorization Predicts}
\label{sec:predictions}
Two consequences follow, and we test both. Each path factor is fit against a decision that every sample in its subtree supervises, whereas a flat softmax resolves the whole $L$-way competition at once; miscalibration should therefore accumulate gradually along a path rather than appear all at once at the leaves, so the gap over flat models should widen with depth. Each ancestor factor gates its entire subtree, so by Prop.~0(iii) no leaf can exceed its ancestor's mass and errors should stay within-subtree. One prediction is negative: evidential shrinkage moves $\hat p_{v,k}$ toward $a_{v,k}$ without reordering samples, so ranking-based uses of confidence such as selective prediction should not improve, and we claim no gain there.

\begin{table*}[t]
\centering
\setlength{\tabcolsep}{5pt}
\caption{Main comparison at leaf-level. bAcc and ECE in \%, NLL in nats; best in \textbf{bold},
second best \underline{underlined}. \emph{Hier.}\ marks a predictor defined over
the label tree, \emph{Unc.}\ one that emits an epistemic uncertainty; only
H\textsuperscript{2}EDL has both, HENN being evidential but flat (its composite
focal sets are a supplied partition, not the taxonomy). An
H\textsuperscript{2}EDL variant ranks first or second in every column, taking
both places in four of the six, and H\textsuperscript{2}EDL is the only method
top-two in bAcc on both datasets; each baseline instead falls away on at least
one axis. The bAcc margins sit inside seed noise, whereas ECE and NLL separate
decisively.
$^\dagger$I-EDL uses its validation-retuned
Fisher weight ($c\!=\!0.02$ at $K\!=\!100$, $c\!=\!0.05$ at $K\!=\!40$); its
published default drops to $16.9\%$ bAcc on FGVC.}
\label{tab:main}
\begin{tabular}{lcc@{\hskip 12pt}ccc@{\hskip 14pt}ccc}
\toprule
& & & \multicolumn{3}{c@{\hskip 14pt}}{FGVC-Aircraft} & \multicolumn{3}{c}{DERM12345} \\
\cmidrule(lr){4-6}\cmidrule(lr){7-9}
Method & Hier. & Unc. & bAcc\,$\uparrow$ & ECE\,$\downarrow$ & NLL\,$\downarrow$ & bAcc\,$\uparrow$ & ECE\,$\downarrow$ & NLL\,$\downarrow$ \\
\midrule
Flat-CE  & \xmark & \xmark & 53.6\sd{0.5} & 19.5\sd{1.9} & 2.00\sd{.10} & 32.9\sd{0.9} & 31.4\sd{1.2} & 2.48\sd{.21} \\
Flat-EDL & \xmark & \cmark & 50.1\sd{2.0} & 10.1\sd{3.9} & 2.36\sd{.10} & 29.3\sd{6.3} & \textbf{14.9}\sd{4.1} & 2.17\sd{.05} \\
Hier-CE  & \cmark & \xmark & 54.2\sd{1.2} & 21.0\sd{1.3} & 2.09\sd{.01} & 32.0\sd{0.4} & 31.7\sd{0.7} & 2.58\sd{.18} \\
HENN     & \xmark & \cmark & 54.2\sd{0.8} & 8.6\sd{2.8} & 1.95\sd{.03} & \textbf{35.8}\sd{1.4} & 18.1\sd{2.6} & 1.99\sd{.02} \\
I-EDL$^\dagger$ & \xmark & \cmark & \underline{54.9}\sd{0.7} & 45.6\sd{0.7} & 3.16\sd{.02} & 30.0\sd{1.9} & 32.8\sd{0.7} & 2.36\sd{.02} \\
\midrule
H\textsuperscript{2}EDL     & \cmark & \cmark & \underline{54.9}\sd{1.7} & 8.8\sd{2.3} & 1.95\sd{.06} & 32.1\sd{2.2} & 16.4\sd{2.5} & \underline{1.80}\sd{.04} \\
\quad+\,leaf-path & \cmark & \cmark & 53.2\sd{0.6} & \textbf{7.0}\sd{2.0} & \underline{1.93}\sd{.04} & 33.8\sd{2.9} & 17.3\sd{1.8} & \textbf{1.78}\sd{.02} \\
\quad+\,consistency         & \cmark & \cmark & 53.0\sd{0.4} & 10.1\sd{2.1} & 2.01\sd{.03} & \underline{34.5}\sd{1.8} & \underline{16.0}\sd{3.5} & \textbf{1.78}\sd{.08} \\
\quad+\,leaf-path+base-rate & \cmark & \cmark & \textbf{55.3}\sd{0.3} & \underline{8.5}\sd{0.3} & \textbf{1.91}\sd{.01} & 33.6\sd{5.1} & 19.1\sd{1.9} & 2.09\sd{.09} \\
\bottomrule
\end{tabular}
\end{table*}

\section{Experiments}
\label{sec:experiments}

\subsection{Experimental Setup}
We evaluate on two hierarchies of very different character.
DERM12345~\citep{derm12345} is a long-tailed dermatoscopic benchmark with a
$4$-level diagnostic taxonomy, $40$ leaves and $15$ branching nodes. We use its
official patient-disjoint split and hold out $10\%$ of patients for
validation.
FGVC-Aircraft~\citep{maji2013fgvc} is a $3$-level fine-grained benchmark with
$30$ manufacturers, $70$ families and $100$ variants on
its official split. All models share an ImageNet-pretrained
ResNet-50~\citep{he2016resnet} and an identical budget of $100$ AdamW
epochs~\citep{loshchilov2019adamw}, are selected on validation leaf balanced
accuracy, and are scored once on test. We report mean\,$\pm$\,std over $3$
seeds with paired $t$-tests, and coarse-level metrics always aggregate leaf
probabilities through Eq.~\eqref{eq:subtree}. The baselines share the backbone
and the recipe. Flat-CE and Flat-EDL~\citep{sensoy2018edl} are the two flat
references. Hier-CE is the softmax twin of H\textsuperscript{2}EDL, with one
local softmax per branching node and the same path product, which isolates the
evidential machinery from the hierarchy itself. HENN~\citep{li2024henn} is a flat
Group-Dirichlet hyper-evidential model over singletons plus a disjoint composite
partition. I-EDL~\citep{deng2023iedl} is a popular flat-EDL variant.

\paragraph{Overview.} The evaluation proceeds in five steps, in the order of the
subsections below.
\textbf{(i)~Classification accuracy} scores the
predictor at the leaf and at the intermediate levels.
\textbf{(ii)~Calibration} and
\textbf{(iii)~mistake severity} test the two positive
predictions of the previous section, and carry most of the empirical
weight.
\textbf{(iv)~Partial supervision} exercises the
setting the path parameterization was built for, in which labels stop at an
interior node.
\textbf{(v)~An ablation} isolates the contribution of
each component and of the baselines' published constants.

\subsection{Hierarchical Classification Accuracy}
\label{sec:accuracy}
To demonstrate that a set of local evidential opinions yields an effective
hierarchical classifier, we compare against the five baselines at the leaf level
(Table~\ref{tab:main}) and at the intermediate levels of the DERM12345 taxonomy
(Table~\ref{tab:derm-coarse}).

As shown in Table~\ref{tab:main}, at the leaf level no method separates itself by a significant margin, but H\textsuperscript{2}EDL is the only method that is
top-two in bAcc on {both} datasets, and an H\textsuperscript{2}EDL variant
places first or second in every one of the six columns of Table~\ref{tab:main},
holding both places in four of them.
Each baseline instead has an axis on which
it falls away: HENN leads DERM bAcc but is fourth on FGVC, I-EDL ties for
second on FGVC bAcc at $45.6\%$ ECE, and Flat-EDL pairs the lowest DERM ECE with
the lowest accuracy of any method.
On FGVC-Aircraft H\textsuperscript{2}EDL+leaf-path+base-rate reaches
$55.3\%$ bAcc, nominally $+1.7$ over Flat-CE and $+1.1$ over Hier-CE, with a
sixth of the seed variance of the plain model ($\pm0.3$ against $\pm1.7$), so
the base-rate prior is stabilizing training and not only shifting the mean. HENN
is nonetheless indistinguishable from either plain H\textsuperscript{2}EDL
($p\!=\!0.47$) or Flat-CE ($p\!=\!0.27$). On DERM12345 HENN attains the highest bAcc
($35.8\pm1.4$), exceeding plain H\textsuperscript{2}EDL ($p\!=\!0.021$) though
not Flat-CE ($p\!=\!0.062$), and our best variant at $34.5\%$ stays tied with
Flat-CE ($p\!=\!0.10$). Two facts here are worth naming. The softmax hierarchy
alone buys nothing: Hier-CE reaches $32.0$ against Flat-CE's $32.9$. And
on the shallow, near-balanced FGVC taxonomy a well-tuned flat evidential head is
already an excellent recognizer: retuned I-EDL matches our plain model at $54.9$.
This parity is a property of the benchmarks rather than
of any method: forcing the ground-truth branching decision down to depth $d$ and
re-taking the argmax within the surviving subtree lifts DERM12345 leaf accuracy
from ${\sim}0.57$ to ${\sim}0.79$ the moment the super-class decision is
supplied, while at any fixed depth the methods differ by at most $0.013$. One
$4$-way decision commits $70$--$74\%$ of all leaf errors, so leaf accuracy on
these benchmarks is largely a proxy for a coarse routing decision that no
fine-level machinery moves.

\begin{table}[t]
\centering
\small
\setlength{\tabcolsep}{4pt}
\caption{Balanced accuracy (\%) at the intermediate levels of the
DERM12345 taxonomy (same runs as Table~\ref{tab:main}). Every
H\textsuperscript{2}EDL variant is above every baseline at Main-1 and Main-2;
leaf-level bAcc, where H\textsuperscript{2}EDL does not lead, is in
Table~\ref{tab:main}. Best in \textbf{bold}, second best \underline{underlined}.}
\label{tab:derm-coarse}
\begin{tabular}{lccc}
\toprule
Method & Super & Main-1 & Main-2 \\
\midrule
Flat-CE  & 86.8\sd{1.9} & 55.5\sd{1.5} & 45.1\sd{1.0} \\
Flat-EDL & 84.9\sd{1.1} & 57.7\sd{5.2} & 46.6\sd{2.8} \\
Hier-CE  & 87.0\sd{1.2} & 55.2\sd{2.7} & 44.9\sd{1.0} \\
HENN     & 87.9\sd{0.7} & 57.3\sd{1.7} & 46.6\sd{0.9} \\
I-EDL    & 74.0\sd{1.1} & 47.9\sd{1.0} & 41.0\sd{0.7} \\
\midrule
H\textsuperscript{2}EDL     & 88.0\sd{0.5} & \underline{59.3}\sd{1.2} & 48.1\sd{0.6} \\
\quad+\,leaf-path           & \textbf{88.6}\sd{0.3} & 59.2\sd{2.4} & \underline{48.3}\sd{2.1} \\
\quad+\,consistency         & 86.7\sd{0.2} & \textbf{59.7}\sd{0.7} & \textbf{48.8}\sd{0.6} \\
\quad+\,leaf-path+base-rate & \underline{88.2}\sd{0.5} & 58.7\sd{0.7} & 48.2\sd{0.8} \\
\bottomrule
\end{tabular}
\end{table}

Scored where the factorization actually operates
(Table~\ref{tab:derm-coarse}), DERM12345 separates cleanly: every
H\textsuperscript{2}EDL variant beats every baseline at both intermediate levels. The consistency variant leads Main-2 at
$48.8\%$ against the best baseline's $46.6$ and Main-1 at $59.7$ against $57.7$, with $+2.2$ and $+2.0$ over the strongest competitor at each level, $+3.7$ and
$+4.2$ over Flat-CE, clearing $p\!<\!0.05$ against Flat-CE ($p\!=\!0.028$ and
$0.012$), Hier-CE ($p\!=\!0.005$) and I-EDL ($p\!=\!0.002$). The two remaining
comparisons are limited by the baselines' own instability: HENN trails by the same $2.2$ at $p\!=\!0.082$, and Flat-EDL's seed
variance at these levels is four to seven times ours ($\pm5.2$ against $\pm0.7$
at Main-1), so an equal gap over it cannot be resolved. Consistency across seeds
is itself part of what the factorization buys. At the super-class level three of
four variants still top every baseline ($88.6$, $88.2$ and $88.0$ against HENN's
$87.9$), on a level where everything above I-EDL already sits near $88\%$ and the
margins are correspondingly inside seed noise ($p\!\ge\!0.12$).

On FGVC-Aircraft we remain ahead at both coarse levels ($78.0$ against Flat-CE's
$77.1$ at manufacturer, $65.3$ against $64.0$ at family).

\subsection{Calibration Across Hierarchy Levels}
\label{sec:calibration}
To test the first prediction of the factorization, that the advantage over flat
models has a specific shape rather than merely an existence, we
report ECE at every level of both taxonomies, and under a doubled training
budget. Each path factor is
supervised by every sample in its subtree, whereas a flat softmax resolves the
whole $L$-way competition at once. The gap should therefore be small where the
decision is easy and well populated, and should grow as the tree deepens and the
evidence thins.

Table~\ref{tab:budget} puts the two models side by side on FGVC-Aircraft at both
budgets. H\textsuperscript{2}EDL is better calibrated at the leaf level by
$2.2\times$ at the standard schedule ($8.8$ against $19.5\%$ ECE) and by
$2.8\times$ when it is doubled ($8.6$ against $24.1$, $p\!=\!0.009$), with the
same ordering on NLL ($1.95$ against $2.00$, then $1.99$ against $2.25$,
$p\!=\!0.006$). The extra budget buys neither method accuracy, but Flat-CE
converts it into $4.6$ points of ECE and $0.25$ nats, while
H\textsuperscript{2}EDL moves by $-0.2$ and $+0.04$: the gap widens with training
rather than closing. This matters for durability. An advantage that shrank as the
baseline trained longer would be an artifact of an under-trained comparison,
whereas one that grows reflects a structural difference in what the two
objectives do with extra capacity to fit. The vacuity term gives each local
decision a data-dependent floor on how confident it may become; cross-entropy has
no such floor and spends the budget sharpening leaf logits it already gets right.

\begin{table}[t]
\centering
\small
\setlength{\tabcolsep}{4pt}
\caption{Calibration under a doubled training budget (FGVC-Aircraft; paired
$t$-test on the $200$-epoch pair). The extra budget buys neither method
accuracy, but Flat-CE converts it into overconfidence and
H\textsuperscript{2}EDL does not, so the gap widens with training.}
\label{tab:budget}
\begin{tabular}{llccc}
\toprule
 & & $100$ ep & $200$ ep & $\Delta$ \\
\midrule
\multirow{2}{*}{bAcc\,$\uparrow$} & Flat-CE & 53.6\sd{0.5} & 53.1\sd{0.7} & $-0.5$ \\
 & H\textsuperscript{2}EDL & 54.9\sd{1.7} & 52.9\sd{1.2} & $-2.0$ \\
\midrule
\multirow{3}{*}{ECE\,$\downarrow$} & Flat-CE & 19.5\sd{1.9} & 24.1\sd{1.8} & $+4.6$ \\
 & H\textsuperscript{2}EDL & \textbf{8.8}\sd{2.3} & \textbf{8.6}\sd{1.7} & $-0.2$ \\
 & \emph{ratio} & $2.2\times$ & $\mathbf{2.8\times}$ & $p\!=\!0.009$ \\
\midrule
\multirow{3}{*}{NLL\,$\downarrow$} & Flat-CE & 2.00\sd{.10} & 2.25\sd{.05} & $+0.25$ \\
 & H\textsuperscript{2}EDL & \textbf{1.95}\sd{.06} & \textbf{1.99}\sd{.04} & $+0.04$ \\
 & \emph{gap} & $0.05$ & $\mathbf{0.26}$ & $p\!=\!0.006$ \\
\bottomrule
\end{tabular}
\end{table}

That advantage resolves into a depth profile rather than an offset. At the coarsest level of either tree,
where the decision is easy and well populated, the two models are statistically
indistinguishable ($p\!\ge\!0.20$ on FGVC-Aircraft), and they separate only as
the evidence thins: $2.8\times$ at FGVC's finest level (Table~\ref{tab:budget}),
and on DERM12345 a leaf ECE of $31.4$ and $31.7\%$ for Flat-CE and Hier-CE
against our $16.4$ (Table~\ref{tab:main}). Every variant roughly halves the CE
models' leaf ECE on that dataset and cuts their NLL by about $30\%$, for all
three seeds. A uniform advantage, or any constant post-hoc temperature, would
have shown up at all levels alike.

Against the flat evidential baselines the result is carried by the proper
scoring rule. H\textsuperscript{2}EDL has the best leaf NLL on both datasets
($1.78$ against HENN's $1.99$ and Flat-EDL's $2.17$ on DERM12345, $1.91$ against
$1.95$ on FGVC-Aircraft), and all four variants beat all five baselines on
super-class ECE. At the two intermediate levels, Flat-EDL and HENN are better calibrated. On leaf ECE Flat-EDL is
lower, $14.9$ against our $16.0$. That comparison should be read with what it
costs: ECE is minimized by never becoming confident, and Flat-EDL buys its
calibration at $29.3\%$ leaf bAcc, the lowest of any method here, carrying the
largest seed variance in Table~\ref{tab:main} on both axes ($\pm6.3$ and
$\pm4.1$). NLL prices miscalibration and vagueness together, and on it the
ordering reverses by a wide margin. We therefore claim the depth profile against
the cross-entropy baselines and the proper-scoring-rule result against the flat
evidential ones --- and neither flat model supplies a coherent coarse read-out at
any level.

\subsection{Severity of the Remaining Errors}
\label{sec:severity}
To test the second prediction of the factorization, that ancestor gating
confines errors to the correct coarse region, we measure two
quantities, both conditioned on the prediction being wrong so that neither
restates accuracy: the depth of the first incorrect branching decision, and the
share of errors that still retain the correct ancestor at a given
level~\citep{bertinetto2020making}.

The motivation is that leaf accuracy treats every error identically and a
taxonomy does not. Confusing two
nevus subtypes and calling a melanoma benign are the same event to a $0/1$ loss
and quite different events in a clinic. Since the number of errors is
fixed by a coarse decision no method improves, what is left to differ is the
kind.

\begin{table}[t]
\centering
\small
\setlength{\tabcolsep}{3.5pt}
\caption{Mistake severity, conditioned on the prediction being wrong
(paired $t$-test vs.\ Flat-CE). Leaf accuracy is statistically tied on
both datasets ($p\!=\!0.21$\,/\,$p\!=\!0.73$), so these differences are in the
kind of error, not the amount.}
\label{tab:severity}
\resizebox{\columnwidth}{!}
{
\begin{tabular}{llccc}
\toprule
& & Flat-CE & H\textsuperscript{2}EDL & $p$ \\
\midrule
\multirow{2}{*}{DERM} & mean first-error depth & 1.207\sd{.028} & \textbf{1.277}\sd{.035} & \textbf{0.003} \\
 & keeps correct main-1 & 25.6\sd{2.5} & \textbf{30.0}\sd{2.5} & \textbf{0.022} \\
\midrule
\multirow{2}{*}{FGVC} & mean first-error depth & 0.590\sd{.018} & \textbf{0.713}\sd{.031} & \textbf{0.036} \\
 & keeps correct manufacturer & 38.9\sd{0.8} & \textbf{46.2}\sd{2.6} & 0.060 \\
\bottomrule
\end{tabular}
}
\end{table}

The effect (Table~\ref{tab:severity}) holds on both datasets, and it is larger on
the bigger label set. On FGVC-Aircraft, $46.2\%$ of H\textsuperscript{2}EDL's
errors still name the correct manufacturer against $38.9\%$ for Flat-CE, a $19\%$
relative improvement in how recoverable its mistakes are, at indistinguishable
leaf accuracy. This is the mechanism of the factorization showing up in the
errors. Each ancestor factor gates its whole subtree at once, so a leaf outside
the plausible region has to overcome a penalty at every level it crosses. A flat
softmax couples nothing, and its objective contains no term that distinguishes a
near miss from a distant one, so the wider the label set the more room its argmax
has to wander out of the correct coarse concept.

\subsection{Learning from Partially Labeled Hierarchies}
\label{sec:partial-main}
To demonstrate that the path parameterization absorbs coarse annotations without
modification, we coarsen a fraction $f$ of DERM12345 training labels to
super-class only and compare against the two standard flat fallbacks.

Everything so far assumes every training image carries a leaf. Dermatology
archives do not work that way: a case confirmed as melanocytic but never subtyped
is a real annotation, not a missing one. Supervision enters Eq.~\eqref{eq:path}
as a set of decisions, so a label naming only an ancestor $v^\star$
shortens the sum instead of degrading it, and the heads below stay free to
express vacuity. No architectural change is involved. We choose the coarsened
subset by a hash
of each sample's stable identifier so that it is identical across methods, seeds
and fractions and nested as $f$ grows. Any difference is then attributable to how
a coarse label is used, not to which samples were coarsened. A flat model has two
standard options and we run both: \emph{soft}, training on the uniform
distribution over $v^\star$'s descendants, and \emph{drop}, discarding the sample.

\begin{table}[t]
\centering
\small
\setlength{\tabcolsep}{3.2pt}
\caption{Coarse-label supervision on DERM12345 ($40$-epoch budget). $f$ is
the fraction of training labels coarsened to super-class only.
H\textsuperscript{2}EDL is ahead on every metric already at $f\!=\!0$ and stays
ahead as labels are coarsened; the quantity of interest is each method's
degradation from its own full-supervision score, so read down each column. Best
in \textbf{bold}.}
\label{tab:partial-main}
\resizebox{\columnwidth}{!}
{
\begin{tabular}{llcccc}
\toprule
& & $f\!=\!0$ & $0.25$ & $0.50$ & $0.75$ \\
\midrule
\multirow{3}{*}{super bAcc\,$\uparrow$}
 & Flat-CE soft & 86.6\sd{1.9} & 83.4\sd{1.5} & 84.7\sd{1.9} & 82.9\sd{3.7} \\
 & Flat-CE drop & 86.6\sd{1.9} & 84.3\sd{1.4} & 85.3\sd{2.9} & 82.8\sd{3.6} \\
 & H\textsuperscript{2}EDL & \textbf{87.4}\sd{1.4} & \textbf{88.0}\sd{0.6} & \textbf{86.2}\sd{1.5} & \textbf{87.0}\sd{1.9} \\
\midrule
\multirow{3}{*}{leaf Acc\,$\uparrow$}
 & Flat-CE soft & 55.8\sd{1.3} & 51.4\sd{3.7} & 45.3\sd{2.5} & 37.6\sd{4.8} \\
 & Flat-CE drop & 55.8\sd{1.3} & 55.3\sd{0.8} & 55.7\sd{1.2} & \textbf{52.0}\sd{1.0} \\
 & H\textsuperscript{2}EDL & \textbf{56.2}\sd{0.5} & \textbf{56.3}\sd{1.5} & \textbf{55.7}\sd{2.1} & 51.2\sd{3.1} \\
\midrule
\multirow{3}{*}{leaf ECE\,$\downarrow$}
 & Flat-CE soft & 30.7\sd{1.0} & 17.5\sd{2.5} & 16.6\sd{0.5} & 20.5\sd{1.8} \\
 & Flat-CE drop & 30.7\sd{1.0} & 27.5\sd{6.4} & 29.8\sd{0.6} & 28.5\sd{1.9} \\
 & H\textsuperscript{2}EDL & \textbf{12.1}\sd{4.0} & \textbf{12.8}\sd{3.7} & \textbf{12.1}\sd{1.0} & \textbf{7.2}\sd{2.6} \\
\midrule
\multirow{3}{*}{leaf NLL\,$\downarrow$}
 & Flat-CE soft & 2.26\sd{.10} & 2.14\sd{.18} & 2.41\sd{.14} & 2.82\sd{.25} \\
 & Flat-CE drop & 2.26\sd{.10} & 2.15\sd{.49} & 2.31\sd{.02} & 2.24\sd{.21} \\
 & H\textsuperscript{2}EDL & \textbf{1.75}\sd{.03} & \textbf{1.74}\sd{.05} & \textbf{1.79}\sd{.05} & \textbf{1.88}\sd{.10} \\
\bottomrule
\end{tabular}
}
\end{table}

The effect appears where the parameterization says it should, at the level the
retained prefix supervises (Table~\ref{tab:partial-main}). Coarsening three
quarters of the labels costs H\textsuperscript{2}EDL $0.4$ points of super-class
balanced accuracy ($87.4\rightarrow87.0$) against $3.7$ for soft and $3.8$ for
drop. A coarse annotation is a complete label for the super-class
decision, and only a model whose supervision is a path of decisions can spend it
that way. Against soft, the standard way a flat model absorbs a coarse label, we
win every metric at every $f\!\ge\!0.25$ and the gap grows with $f$: at
$f\!=\!0.75$ we hold $51.2\%$ leaf accuracy against $37.6\%$, having lost $5.0$
points from our own full-supervision score where soft lost $18.2$. Fabricating a
uniform target over $v^\star$'s descendants corrupts the fine-grained heads, and
the damage compounds. Calibration is the most robust effect: leaf ECE never
exceeds $12.8\%$ and improves to $7.2\%$ at $f\!=\!0.75$ as vacuity
absorbs the supervision that is no longer there, while drop never falls
below $27.5\%$ ($p\!=\!0.002$).

Drop is the harder baseline and we do not beat it on accuracy. It matches us on
leaf accuracy at every fraction and leads at $f\!=\!0.75$ ($52.0$ against
$51.2$). Training on a quarter of the leaf labels is evidently
close to training on all of them at this scale, which caps what any
hierarchy-native use of the discarded labels can add. What drop cannot do is stay
calibrated, holding $28.5\%$ leaf ECE and $2.24$ nats where we hold $7.2\%$ and
$1.88$. The narrow claim these numbers support is that a coarse annotation is a
complete label for the level it names: coarsening three quarters of the leaves
costs H\textsuperscript{2}EDL $0.4$ points of super-class accuracy against $3.7$
for the flat fallback that keeps those samples, while the fallback that matches
our accuracy does so only by discarding them and never falls below $27.5\%$ leaf
ECE where we reach $7.2$.

\subsection{Ablation Study and Hyperparameter Transfer}
\label{sec:ablation}
To assess how the individual components contribute, we read
Table~\ref{tab:main} as a component ablation, and then examine how each
baseline's published constants transfer across class counts.

Which components help
turns out to be dataset-dependent in an informative way. On FGVC-Aircraft
the base-rate prior is the single most effective addition, giving the best
accuracy, the best calibration and most of the variance reduction. On DERM12345
it does not transfer. There it raises seed variance (bAcc $\pm5.1$ against
$\pm1.8$--$2.9$) and worsens both ECE and NLL, which leaves the parameter-free
consistency variant as the strongest DERM configuration. The mirror image
holds: on FGVC-Aircraft that same consistency term is the weakest
H\textsuperscript{2}EDL variant ($53.0\pm0.4$ bAcc, ECE $10.1\pm2.1$), below the
plain model on all three metrics and, notably, not even more path-consistent at
test time ($0.923$ against $0.931$ without it; \emph{path consistency} is the
label-free fraction of test samples whose argmax at every level agrees with the
leaf argmax projected to that level); neither component is a free addition. We attribute the base-rate reversal to DERM's short, heavily
long-tailed frame, in which a tempered inverse-frequency prior over-corrects the
rare leaves. End-to-end leaf coupling is
a mild but reliable gain on both datasets.

A parallel lesson applies to the flat evidential references. Flat-EDL and HENN
transfer with no retuning at all. I-EDL's Fisher weight,
by contrast, must scale with the class count: its published $c\!=\!0.05$ drives
all evidence to zero at $K\!=\!100$, leaving $16.9\%$ bAcc, and needs
$c\!=\!0.02$ before it becomes competitive. This is a systematic consequence of the class count. Alongside our own base-rate result it is
a second reminder that constants published for one class count do not port to a
$40$- or $100$-way fine-grained tree. The ablation therefore points one way: most
of what the hierarchy has to offer is already supplied by the path factorization
itself, the base-rate prior and the consistency term each help on exactly one of
the two benchmarks, and neither they nor a baseline's published constant survives
a change of class count unchanged.

\section{Conclusion}
H\textsuperscript{2}EDL makes hyper-evidential uncertainty tractable on
tree-structured label spaces, by observing that the taxonomy is already the right
hyperdomain. Its subtrees and leaf singletons form a linear-size focal family on
which the belief masses are induced in closed form, by one ordinary
Dirichlet opinion per branching decision, instead of being predicted. What comes
out is a single parameter set with two readings: an operational classifier with
exact coarse marginals, and a valid tree-structured hyper-opinion whose composite
mass at a node is the belief that reached it and declined to descend, with a
path-local imprecision band (Prop.~2) and a per-sample depth profile $\{U_d\}$
(Prop.~3) that a flat evidential model cannot express. No second
network is required, nor composite annotations, nor a post-hoc calibration stage.
Empirically, leaf accuracy on both benchmarks is dominated by one coarse decision
that no method improves, so parity there says more about the data than about the
model. Scored at the levels the factorization governs, every variant leads all
five baselines at both intermediate DERM12345 levels, the ECE of the cross-entropy
baselines is roughly halved by a margin that grows with depth and with training
budget, and the remaining errors stay inside the correct coarse category far more
often at indistinguishable accuracy. Of these the depth profile is the stronger
test, since a constant post-hoc rescaling could reproduce the average but not the
shape.

\bibliography{h2edl}

\end{document}